\documentclass{bmvc2k}

\title{ContourletNet: A Generalized Rain Removal Architecture Using Multi-Direction Hierarchical Representation}

\addauthor{Wei-Ting Chen}{jimmy3505090@gmail.com}{1,3*}
\addauthor{Cheng-Che Tsai}{r08943148@ntu.edu.tw}{1*}
\addauthor{Hao-Yu Fang}{danielfang60609@gmail.com}{2}
\addauthor{I-Hsiang Chen}{f09921058@ntu.edu.tw}{2}
\addauthor{Jian-Jiun Ding}{jjding@ntu.edu.tw}{2}
\addauthor{Sy-Yen Kuo}{sykuo@ntu.edu.tw}{2}

\addinstitution{
 Graduate Institute of Electronics Engineering\\
 National Taiwan University\\
 Taipei, Taiwan
}
\addinstitution{
 Department of Electrical Engineering\\
 National Taiwan University\\
 Taipei, Taiwan
}
\addinstitution{
 ASUS Intelligent Cloud Services\\
 Asustek Computer Inc.\\
 Taipei, Taiwan}

\runninghead{W.T Chen and C.C Tsai et al.}{Generalized Rain Removal Architecture}

\usepackage[ruled,vlined]{algorithm2e}
\usepackage{amsmath,graphicx}

\usepackage{booktabs,subfigure}
\usepackage{comment}
\usepackage[skip=2pt]{caption}
\newcommand{\tabref}[1]{Table~\ref{#1}}
\newcommand{\figref}[1]{\figurename~\ref{#1}}

\begin{document}

\maketitle

\begin{abstract}
Images acquired from rainy scenes usually suffer from bad visibility which may damage the performance of computer vision applications. The rainy scenarios can be categorized into two classes: moderate rain and heavy rain scenes. Moderate rain scene mainly consists of rain streaks while heavy rain scene contains both rain streaks and the veiling effect (similar to haze). Although existing methods have achieved excellent performance on these two cases individually, it still lacks a general architecture to address both heavy rain and moderate rain scenarios effectively. In this paper, we construct a hierarchical multi-direction representation network by using the contourlet transform (CT) to address both moderate rain and heavy rain scenarios. The CT divides the image into the multi-direction subbands (MS) and the semantic subband (SS). First, the rain streak information is retrieved to the MS based on the multi-orientation property of the CT. Second, a hierarchical architecture is proposed to reconstruct the background information including damaged semantic information and the veiling effect in the SS. Last, the multi-level subband discriminator with the feedback error map is proposed. By this module, all subbands can be well optimized. This is the first architecture that can address both of the two scenarios effectively. The code is available in \url{https://github.com/cctakaet/ContourletNet-BMVC2021}.
\end{abstract}

\begin{figure}
  \centering
  \subfigure[Moderate Rain Input]{\label{fig:blocks_basic}\includegraphics[width=0.245\textwidth]{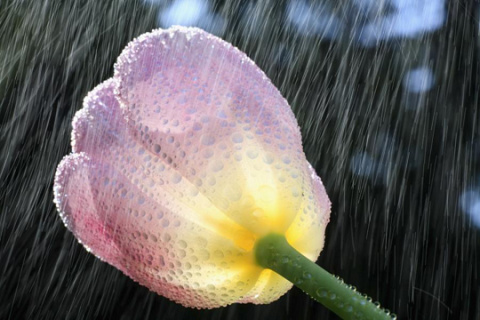}}
  \subfigure[MSPFN\unskip~\cite{jiang2020multi}]{\label{fig:blocks_bottleneck}\includegraphics[width=0.245\textwidth]{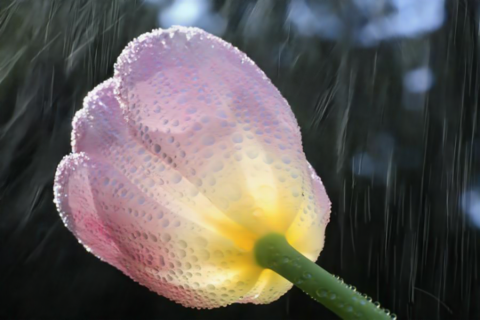}}
  \subfigure[RCDNet\unskip~\cite{wang2020model}]{\label{fig:blocks_bottleneck}\includegraphics[width=0.245\textwidth]{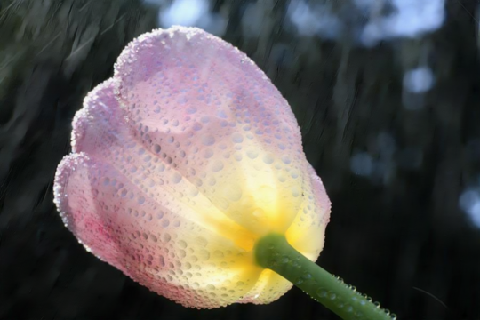}}
  \subfigure[Ours]{\label{fig:blocks_bottleneck}\includegraphics[width=0.245\textwidth]{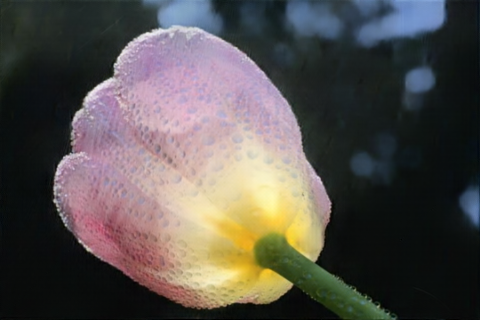}}
  \\
  \subfigure[Heavy Rain Input]{\label{fig:blocks_basic}\includegraphics[width=0.245\textwidth]{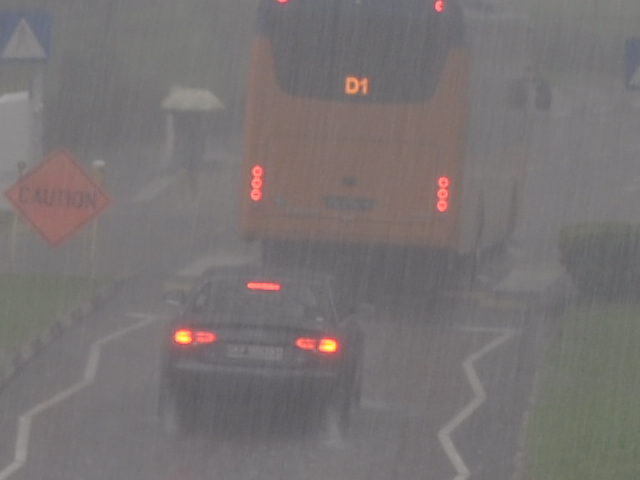}}
  \subfigure[HRGAN\unskip~\cite{li2019heavy}]{\label{fig:blocks_bottleneck}\includegraphics[width=0.245\textwidth]{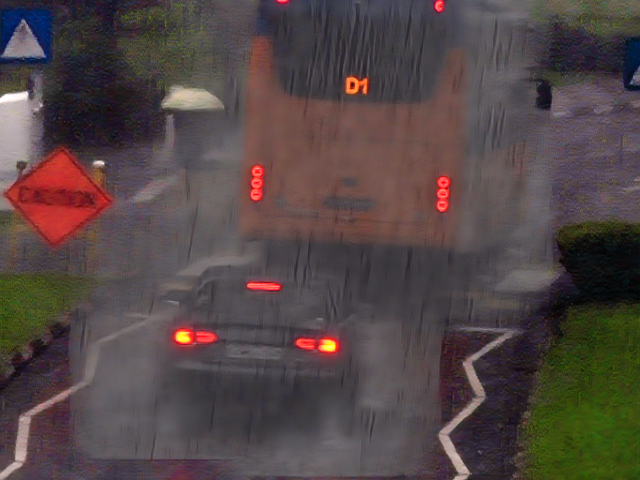}}
  \subfigure[AIO\unskip~\cite{li2020all}]{\label{fig:blocks_bottleneck}\includegraphics[width=0.245\textwidth]{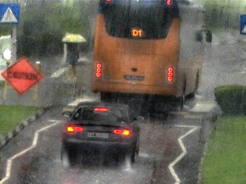}}
  \subfigure[Ours]{\label{fig:blocks_bottleneck}\includegraphics[width=0.245\textwidth]{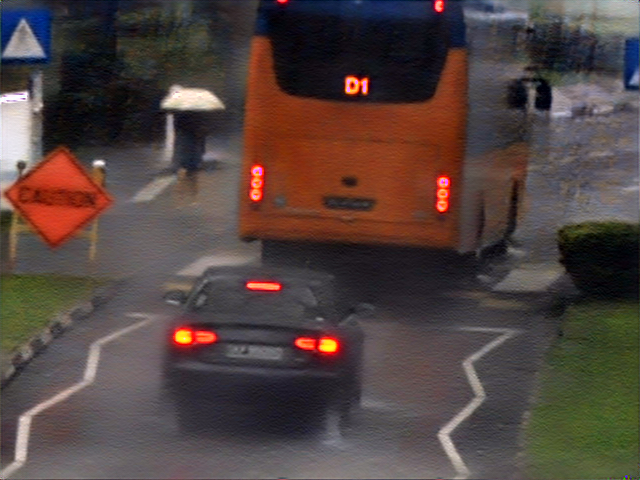}}
  \caption{Visual comparison of the proposed method and state-of-the-art rain removal methods under moderate rain and heavy rain scenarios. It shows that the proposed method has generalized ability for addressing various rain scenarios.}
  \vspace{-0.4cm}
\label{fig:example}
\end{figure}

\section{Introduction}
\label{sec:intro}
Rain streaks is an atmospheric phenomenon which usually leads to poor visibility. Similar to haze~\cite{chen2019pms} and snow~\cite{chen2021all}, it may degrade the performance of high-level vision applications, such as object detection and semantic segmentation. Rainy scenarios can be categorized into: (i) the moderate rain scene and (ii) the heavy rain scene (please refer \figref{fig:example}). The moderate rain scene mainly consists of rain streaks with irregular distribution and can be modeled as:
\begin{equation}
\setlength\abovedisplayskip{0pt}
\mathbf{I}=\mathbf{{J}}+\sum_i^{n}{\mathbf{S}_i},
\setlength\abovedisplayskip{0pt}
\end{equation}
where $\mathbf{S}_i$ is the rain streak in the $i^{th}$ layer, \textbf{I} is the rain image, and \textbf{J} is the clean image. Several algorithms have been proposed to address the moderate rain scenario\unskip~\cite{yang2020wavelet,kang2011automatic,jiang2017novel,fu2017removing,yang2019scale,li2016rain,wei2017should,wang2019spatial,yasarla2020syn2real,jiang2020multi,deng2020detail,wang2020model,du2020variational}. Kang\textit{ et al.}\unskip~\cite{kang2011automatic} applied the image decomposition technique based on dictionary learning and morphology. Fu\textit{ et al.}\unskip~\cite{fu2017removing} adopted discriminatively intrinsic characteristics, Li\textit{ et al.}\unskip~\cite{li2016rain} applied layer separation, and Wang\textit{ et al.}\unskip~\cite{wang2019spatial} proposed the SPANet to remove rain streaks. Jiang\textit{ et al.}\unskip~\cite{jiang2020multi} proposed the multi-scale progressive fusion strategy to achieve excellent results in moderate rain removal.

Unlike moderate rain scenarios, heavy rain scenes usually contain rain streaks and strong veiling effect which is similar to the haze. Its formulation can be expressed as\unskip~\cite{li2019heavy}:
\begin{equation}
\setlength\abovedisplayskip{0pt}
\mathbf{I}=\mathbf{T}\odot(\mathbf{{J}}+\sum_i^{n}{\mathbf{S}_i})+(\mathbf{1}-\mathbf{T})\odot\mathbf{{A}},
\setlength\abovedisplayskip{0pt}
\label{eq:infer}
\end{equation} 
where \textbf{A} is the atmospheric light, \textbf{T} is the transmission map, and $\odot$ is pixel-wise multiplication. Due to multi-flux scattering and veiling effect, the background information under the heavy rain scenario is blurrier than that in the moderate rain case. Li \textit{et al.}\unskip~\cite{li2019heavy} proposed a two-stage optimization strategy termed the HRGAN which extracts \textbf{T}, \textbf{A}, and \textbf{S} for physical model-based recovery, and applied adversarial learning to optimize the results. 

\begin{figure}[t!]
  \centering
  \subfigure[Input]{\includegraphics[width=0.195\textwidth]{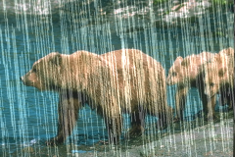}}
  \subfigure[HRGAN]{\includegraphics[width=0.195\textwidth]{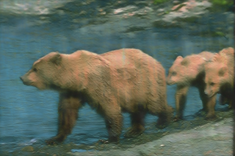}}
  \subfigure[MSPFN]{\includegraphics[width=0.195\textwidth]{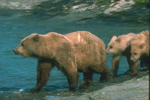}}
  \subfigure[Ours]{\includegraphics[width=0.195\textwidth]{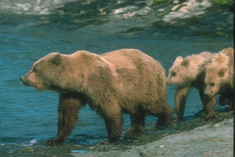}}
  \\
  \subfigure[Input]{\includegraphics[width=0.195\textwidth]{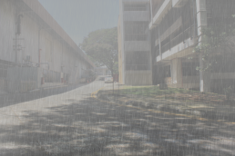}}
  \subfigure[MSPFN]{\includegraphics[width=0.195\textwidth]{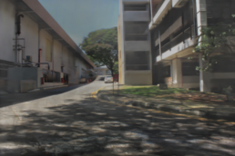}}
  \subfigure[MSPFN+Dehaze]{\includegraphics[width=0.195\textwidth]{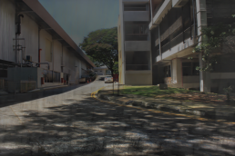}}
  \subfigure[HRGAN]{\includegraphics[width=0.195\textwidth]{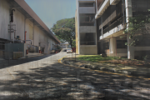}}
  \subfigure[Ours]{\includegraphics[width=0.195\textwidth]{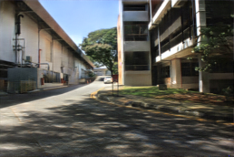}}
\caption{\textbf{Illustration of the limitation in existing methods.} (a) and (e): the inputs of moderate rain and heavy rain. (b)-(d): derained results by the heavy rain removal method (i.e., HRGAN\unskip~\cite{li2019heavy}), MSPFN\unskip~\cite{jiang2020multi} and the proposed method. (f): derained results by moderate rain removal (i.e., MSPFN). (g): MSPFN + dehazed method\unskip~\cite{Dong_2020_CVPR}. (h): HRGAN and (i): the proposed method. Note that, for the fair comparison, both (b) and (f) are retrained with moderate rain and heavy rain (HR) datasets, respectively.}
\label{fig:fig2}
\end{figure} 

Although existing rain removal methods can reconstruct desirable results for the moderate rain or the heavy rain scenario individually, there exists a limitation. These methods cannot address both types of rain scenes with the same architecture. In \figref{fig:fig2}, we present an example that applies a heavy rain removal method (the HRGAN\unskip~\cite{li2019heavy}) to address moderate rain scenes and uses a moderate rain removal method (the MSPFN\unskip~\cite{jiang2020multi}) to handle the heavy rain scenes. We can observe that, for the former case, the recovered results tend to have color distortion problem because the veiling effect removal in the HRGAN may over-dehaze an moderate rain image. Moreover, the result tends to have residual rain streaks. For the latter case, the residual veiling effect may exist in the recovered results because the veiling effect removal is not considered by the moderate rain removal. Moreover, even if we combine the state-of-the-art veiling effect removal strategy (i.e., dehazing strategy)\unskip~\cite{Dong_2020_CVPR} with moderate rain removal, the recovered results are still limited because the heavy rain scenario is not a simple combination of rain streaks and the veiling effect. It suffers from the multi-flux scattering\unskip~\cite{li2019heavy}, resulting in blurry problem in the heavy rain scene.

Thus, in this paper, to adequately address both moderate rain and heavy rain scenarios with one architecture, we propose an effective recovered framework called the ContourletNet. First, the contourlet transform (CT)\unskip~\cite{do2005contourlet} is embedded in our network to generate several multi-direction subbands (MS) and one semantic subband (SS). The MS contains the information of rain streaks and detail such as edges and textures while the SS mainly consists of the contextual information and the veiling effect. With the CT, the information of rain streaks can be well retrieved because rain streaks are usually in different directions. We construct two sub-networks termed multi-direction subband recovery (MSR) and semantic subband recovery (SSR) to recover all subbands adequately. Second, to achieve better recovery of the SS, we investigate both heavy rain and moderate rain images. We find an interesting phenomenon: with the increase of the CT decomposition level, the difference between heavy rain/haze images and the corresponding ground truths in the SS is reduced. Based on this observation, we propose to integrate the hierarchical architecture and the CT in our network. Last, to improve the performance of the ContourletNet, the multi-level semantic subbands discriminator with the feedback error map is proposed. All semantic subbands and the recovered image are distinguished by the multi-level discriminator to enhance the contextual quality.

Experiments show that our proposed method achieves superior performance than state-of-the-art approaches for both moderate rain and heavy rain scenarios. As far as we know, our method is the first work which can deal with moderate rain and heavy rain with one architecture effectively. 
\section{Proposed Method}
\subsection{Effective rain streak extraction}
Rain streak extraction is essential for moderate rain and heavy rain scenarios. We first introduce the contourlet transform (CT) and illustrate why it is helpful for rain streak extraction.
\noindent \smallskip\\
\textbf{Contourlet Transform.} The CT\unskip~\cite{do2005contourlet} is an effective technique for geometric information analysis. It can achieve better representation in both locality and directionality. Its detail is presented in Supplementary Material. It mainly consists of two operations: the Laplacian pyramid (LP)\unskip~\cite{burt1983laplacian} and directional filter banks (DFB)\unskip~\cite{bamberger1992filter}. Given an image, first, the LP filter decomposes it into a semantic subband (SS) and a high-frequency subband. Then, the high-frequency subband is further decomposed into several subspaces with $2^{k}$ direction via directional filters. We term these subspaces multi-direction subbands (MSs). With the CT, the multi-resolution and multi-direction features can be extracted effectively.


\begin{figure}[t!]
\centering \makeatletter{\includegraphics[width=1.0\textwidth]{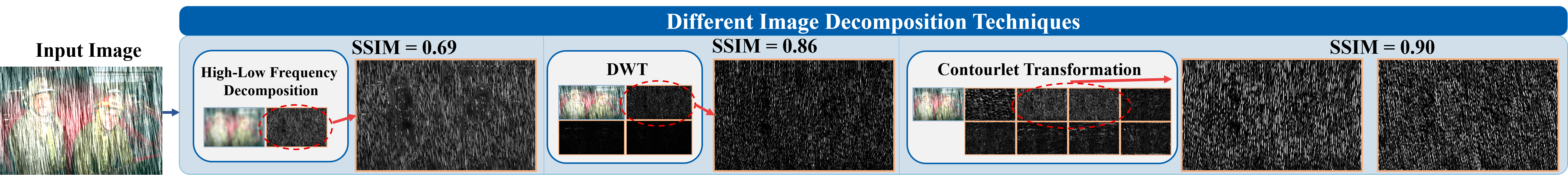}}{}
\makeatother 
\caption{\textbf{Comparison of using different rain streak extraction methods.} One can see that the CT can achieve better rain streaks retrieval, especially for the bevelled directions. The rain streaks of input image mainly contain vertical and bevelled direction. Moreover, the value of SSIM indicates the structural similarity between the rain mask and each subband. Note that we add two subbands to calculate the SSIM for the CT.}
\label{fig:r_streak_compare}
\end{figure}
\noindent \smallskip\\
\textbf{Motivation.} For both moderate rain and heavy rain removal, rain streak extraction plays a crucial role. Rain streaks are not uniform but have different directions and intensities. There are some existing strategies which can extract rain streaks, including high-low frequency decomposition (HL)\unskip~\cite{li2019heavy}, the discrete wavelet transform (DWT)\unskip~\cite{yang2019scale}, and learnable convolution kernels with a guidance\unskip~\cite{wang2020model}. 
For the DWT and HL, as shown in \figref{fig:r_streak_compare}, although they can extract some rain streaks information, the performance is still limited compared with the CT. The reason is as follows. The DWT only contains filter with two directions (i.e., vertical and horizontal), which leads to that the interpretation of bevelled rain streaks is limited. For HL, the information of rain streaks cannot be retrieved appropriately and stably because all high-frequency information (e.g., edges, details, and rain streaks) in various directions are included in a single subband. Moreover, it has no lossless inverse transformation, which limits the performance  of reconstruction. For learnable convolution kernels, though rain information in different directions can be extracted by the well-trained filter from the CNN, achieving global optimization is challenging since the patterns of rain streak in real-world scenarios are too complicated to be covered and modeled comprehensively. Therefore, to address these limitations, we embed the CT into our network to extract rain streaks with better spatial locality and directionality. With this property, the effective representation of rain can be achieved by leveraging the empirical decomposition process based on the prior knowledge of rain streaks.

\subsection{Hierarchical Decomposition for Rain Removal}
\begin{figure}[t!]
\centering 
\includegraphics[scale=0.32]{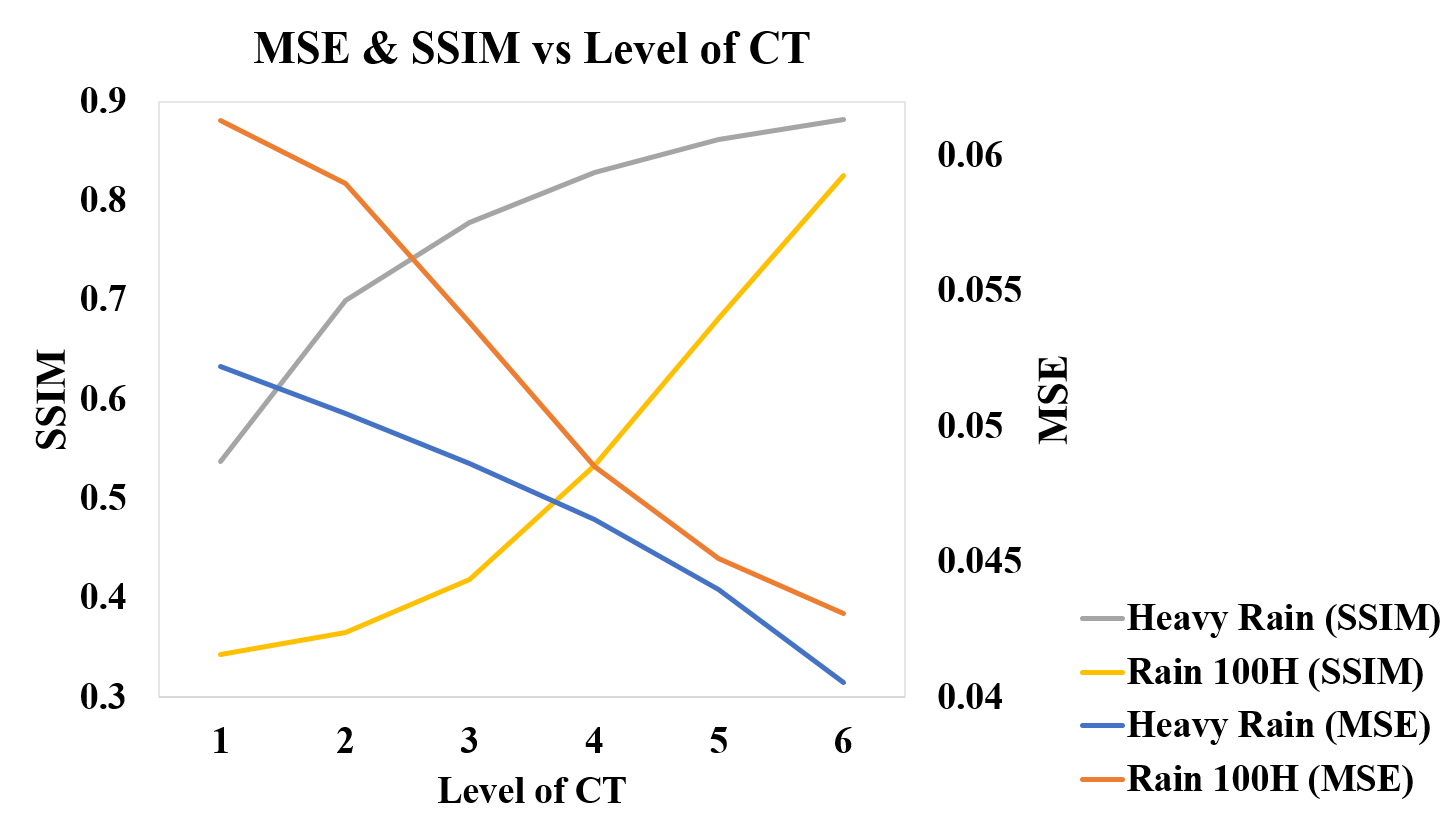}
\caption{\textbf{Investigating the response of the CT level:} The MSE and the SSIM of the rain images (Rain 100H and heavy rain datasets) and the corresponding ground truths in the semantic subband (i.e., SS).}
\label{fig:difference_level}
\end{figure}

The CT decomposes an image into multi-direction subbands (MSs) and the semantic subband (SS). Although appropriate recovery of the MS components can be achieved effectively by the residual strategy due to their sparsity\unskip~\cite{554381:14619341,554381:14619480}, reconstructing the SS component may be challenging since it generally contains complicated content such as residual rain streaks, the degraded background, and the veiling effect. To alleviate this issue, we propose a hierarchical architecture which only decomposes the SS component in each CT level. Specifically, we only recover the SS component at the bottom level and reconstruct several MS components in other levels. This idea is inspired by the investigation on the Rain 100H\unskip~\cite{yang2017deep} and heavy rain datasets\unskip~\cite{li2019heavy} which is shown in \figref{fig:difference_level}. We observed that the difference between the rain image and the corresponding ground truth in the SS may decrease with the number of levels, which means that hierarchical decomposition can benefit the reconstruction of the SS.

For the moderate rain scene, with the hierarchical CT, most rain streaks are retrieved in MS components, which means that the SS component is closer to the ground truth. For the heavy rain scenario, rain streaks can be extracted by MS components, too. However, heavy rain images also contain the veiling effect, which mainly consists of the scene radiance \textbf{J}, the atmospheric light \textbf{A}, and the transmission \textbf{T}\unskip~\cite{li2019heavy}. \textbf{A} is usually assumed as a constant. Although \textbf{T} is a depth-related feature, it may become a constant-like matrix after an adequate iteration of decomposition because high-frequency components such as edge or texture are extracted. Thus, according to the formulation of veiling effect, the relation between the scene radiance \textbf{J} and the heavy rain image \textbf{I} can be approximated to a constant matrix mapping. Based on the above analysis, the difficulty of recovering the SS component can be reduced effectively with the proposed hierarchical CT. Therefore, the combination of the CT and the hierarchical architecture can both improve the ability of feature representation and achieve a better local optima. 

\subsection{Network Architecture}
\begin{figure*}[t!]
\centering \makeatletter{\includegraphics[width=0.95\textwidth]{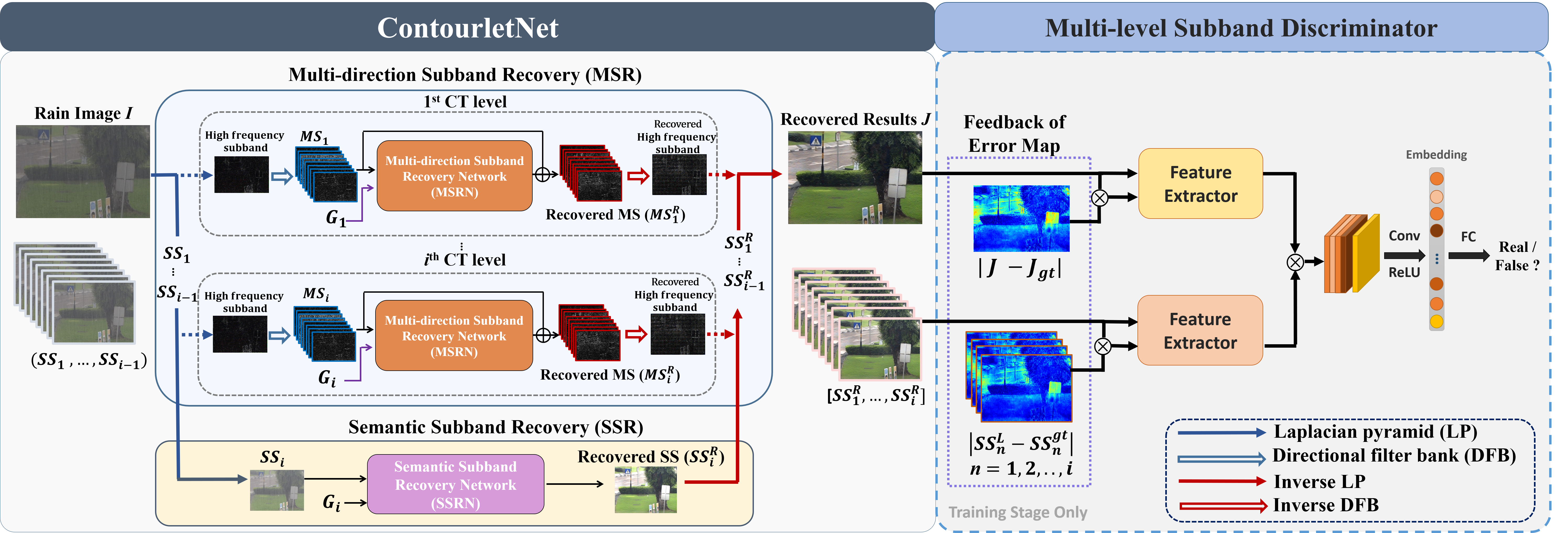}}{}
\makeatother 
\caption{\textbf{The overview of the proposed rain removal network.} It consists of the ContourletNet and the multi-level subband discriminator. The ContourletNet contains two main modules: multi-direction subband recovery (MSR) and semantic subband recovery (SSR). The feature extractor in multi-level subband discriminator is based on Res2Net\unskip~\cite{554381:14619523}.}
\label{fig:network_architecture}
\end{figure*}

The flowchart of the proposed network is shown in \figref{fig:network_architecture}. It consists of two subnetworks: (i) the ContourletNet and (ii) the multi-level subband discriminator. We leverage the GAN\unskip~\cite{620844:14623099} to optimize the performance of the ContourletNet. First, the input image is fed into the ContourletNet to produce the clean image. Then, different from other learning-based architectures\unskip~\cite{zhang2019image, isola2017image}, the recovered image and its recovered MS at different levels are distinguished by the discriminator and the feedback of the error map mechanism. The architecture is illustrated as follows.
\noindent \smallskip\\
\textbf{ContourletNet.} The ContourletNet can be divided into: i) multi-direction subband recovery (MSR) and ii) semantic subband recovery (SSR). First, the CT decomposes the input image into the MS and the SS. Then, the MS is reconstructed by the multi-direction subband recovery network (MSRN). The SS component at the bottom level is recovered by the semantic subband recovery network (SSRN) with pixel-to-pixel reconstruction. Two sub-networks are based on the proposed contourlet predictor (CP) architecture to recover the components. More details about CP, SSRN, and MSRN are presented in the Supplementary Material. Initially, the input is concatenated with the proposed aggregate contourlet component \textit{G\ensuremath{_{i}}\ensuremath{^{}}}:
\begin{equation} 
G_i^{}=\sigma (\mathbf{I})\oplus\sigma (\mathbf{SS_1^{}})\oplus\dots\oplus\sigma (\mathbf{SS}_{i-1}^{}),
\label{eq:g}
\end{equation}
where \textit{G\ensuremath{_{i}}\ensuremath{^{}}} denotes the aggregate contourlet component at the \textit{i}\ensuremath{^{ th}} level, $\oplus$ is the concatenate operation, \textbf{I} is the rainy image, \textbf{SS}\ensuremath{_{i}}\ensuremath{^{}} is the SS component at level \textit{i}, and \textit{\ensuremath{\sigma  } }is the multi-pooling architecture\unskip~\cite{JSTASRChen}. The idea in \eqref{eq:g} is that, instead of down-sampling the input directly, the aggregate contourlet component enables the reconstruction network to acquire more information from the SS components in previous layers. The SS part contains semantic and contextual information which can benefit the reconstruction process. The use of multi-pooling is to prevent the information loss which usually happens in down-sampling. Then, the MSRN and SSRN will reconstruct the information in each subband based on the input and the aggregate contourlet component.
\noindent \smallskip\\
\textbf{Multi-Level subbands Discriminator.} To better optimize the recovered results, the multi-level subbands discriminator is proposed, as shown in the right side of \figref{fig:network_architecture}. In previous works, the final recovered result is generally fed into the discriminator directly. This may limit the performance because it is hard to optimize SS components at all levels based on the final recovered result. To address this, we propose to distinguish both the recovered result and all recovered SS components jointly. Our idea is that the recovered SS is an important information to evaluate the performance of each level because it contains both multi-direction information and semantic information in the previous layers of the inverse CT. With the SS, the information in the previous layer can be optimized simultaneously. It can prevent the error from amplifying massively if the error is generated at the bottom layer. Moreover, to improve the performance of the discriminator, inspired by the attention mechanism, the feedback of the error map is proposed. That is, the input is multiplied by the error map, which is the L1-norm difference between the input and its ground truth. Then, the inputs are concatenated with their corresponding attentive component to the discriminator. It can make the affected regions get more attention from the discriminator.

\subsection{Loss Function}In the proposed hierarchical contourlet-based rain removal network, three losses are applied: (i) contourlet loss, (ii) the perceptual loss, and (iii) the adversarial loss.
\noindent \smallskip\\
\textbf{Contourlet Loss.} The contourlet loss is defined as follows:
\begin{equation} 
\setlength\abovedisplayskip{0pt}
\mathcal{L}_{C}={\left\ensuremath{\Vert}(\mathbf{SS}_{i}^{R}-\mathbf{SS}_{i}^{gt})\right\ensuremath{\Vert}_2}+{\sum\limits_{m=1}^{i}\;}{\sqrt{\left\ensuremath{\Vert}(\mathbf{MS}_{m}^{R}-\mathbf{MS}_{m}^{gt})\right\ensuremath{\Vert}^2+\epsilon^{2}}},
\setlength\abovedisplayskip{0pt}
\end{equation}
where $\mathbf{SS}_{i}^{R}$ and $\mathbf{SS}_{i}^{gt}$ are the predicted semantic subband component and its corresponding ground truth at the $i^{th}$ level. $\mathbf{MS}_{i}^{R}$ and $\mathbf{MS}_{i}^{gt}$ are the predicted multi-direction subbands and their corresponding ground truths, and $\epsilon$ is the slack value which can keep the values in multi-direction subbands non-zero to prevent the texture details from vanishing\unskip~\cite{lai2018fast}.
\noindent \smallskip\\
\textbf{Adversarial Loss.} It is defined as
\begin{equation}
\setlength\abovedisplayskip{0pt}
\mathcal{L}_{Adv}=\underset G{\mbox{min}}\underset D{\mbox{max}}\;[\mathbf{E}_{(\mathbf{{J}_{SS}+J})\sim p(\mathbf{{J}_{SS}+J)}}\left[\log D\left(\mathbf{{J}_{SS}+J}\right)\right]\\
+ \mathbf{E}_{\mathbf{({I}_{SS}+I})\sim p(\mathbf{{I}_{SS}+I)}}\left[\log(1-D(G\left(\mathbf{{I}_{SS}+I}\right))]\right],
\setlength\abovedisplayskip{0pt}
\end{equation}
where $\mathbf{{J}_{SS}+J} $ and $\mathbf{I_{SS}+I} $ are the sets of the clean images and rain images with their corresponding SS components, respectively. D is the discriminator, and G is the generator. The overall loss of the proposed network is
\begin{equation}
\setlength\abovedisplayskip{0pt}
\mathcal{L}_{Overall}=\mathcal{L}_{C}+\lambda_1\mathcal{L}_{Perceptual}+\lambda_2\mathcal{L}_{Adv},
\setlength\abovedisplayskip{0pt}
\end{equation}
where $\mathcal{L}_{Perceptual}$ is the perceptual loss\unskip~\cite{554381:14619567} and we set the $\lambda_{1}=10^{-3}$ and $\lambda_{2}=10^{-4}$.

\section{Experimental Result}
\subsection{Dataset and Implementation Detail}
Several rain datasets are leveraged to validate the derained ability of the proposed method. For the moderate rain, we adopt the Rain100H and Rain100L datasets proposed in\unskip~\cite{yang2017deep}, and Rain800\unskip~\cite{zhang2019image} for training and evaluation.
For the heavy rain, we adopt the heavy rain dataset proposed in\unskip~\cite{li2019heavy}. The details of the datasets can be found in the Supplementary Material.
\begin{table*}[t!]
\caption{\textbf{Quantitative evaluation for comparison with other existing methods on the heavy rain dataset.} The \textbf{DHF} and \textbf{DRF} denote dehaze first and derain first, respectively.}
\centering
\def\arraystretch{1}
\ignorespaces 
\centering 
\scalebox{0.48}{
\begin{tabular}{ccccccccccccccc} 
\toprule
\textbf{Metrics} & \multicolumn{2}{c}{\textbf{RCDNet}\unskip~\cite{wang2020model}} & \multicolumn{2}{c}{\textbf{MSPFN}\unskip~\cite{jiang2020multi}} & \multicolumn{2}{c}{\textbf{BRN}\unskip~\cite{ren2020single}} & 
\textbf{RCDNet*} & 
\textbf{MSPFN*} & 
\textbf{BRN*} & \textbf{CycleGAN}\unskip~\cite{zhu2017unpaired} & 
\textbf{Pix2Pix}\unskip~\cite{isola2017image} & 
\textbf{HRGAN}\unskip~\cite{li2019heavy}&
\textbf{AIO}\unskip~\cite{li2020all}&
\textbf{Ours}
\\ 
\cline{2-7}
 & DHF & DRF & DHF & DRF & DHF & DRF &  &  &  &  &  & & &   \\ 
\hline\hline
\textbf{PSNR}  & 16.42 & 15.81 & 16.69 & 16.03 & 16.76 & 16.68 & 23.15 & 23.92 & 22.38 & 21.62 & 22.43 & 24.78  & 24.71 & \textcolor{red}{{\textbf{25.45}}}  \\ 
\hline
\textbf{SSIM}  & 0.741 & 0.725 & 0.776 & 0.734 & 0.748 & 0.726 & 0.897 & 0.872 & 0.874 & 0.826 & 0.854 & 0.882 & 0.898 & \textcolor{red}{{\textbf{0.912}}} \\ 
\hline
\textbf{CIEDE 2000}  & 13.32 & 14.01 & 12.77 & 13.69 & 12.63 & 13.36 & 5.14 & 4.52 & 6.38 & 7.23 & 6.26 & 7.40 & 4.89 & \textcolor{red}{{\textbf{4.07}}} \\
\bottomrule
\label{tab:quantitative_syn}
\end{tabular}}
\end{table*}

\begin{table*}[t!]
\caption{\textbf{Quantitative evaluation for comparison with other state-of-the-art methods on the existing moderate rain datasets.} (PSNR/SSIM/CIEDE2000)}
\centering
\def\arraystretch{1}
\ignorespaces 
\centering 
\scalebox{0.45}{
\begin{tabular}{cccccccccc} 
\toprule
\textbf{Dataset}  & \textbf{DDN}\unskip~\cite{fu2017removing} & \textbf{JORDER}\unskip~\cite{yang2017deep} & \textbf{SPANet}\unskip~\cite{wang2019spatial} & \textbf{PreNet}\unskip~\cite{ren2019progressive} & \textbf{BRN}\unskip~\cite{ren2020single} & \textbf{DRDNet}\unskip~\cite{deng2020detail} & \textbf{RCDNet}\unskip~\cite{wang2020model} & \textbf{MSPFN}\unskip~\cite{jiang2020multi} & \textbf{Ours}  \\ 
\hline\hline
\textbf{Rain100L} & 32.17/0.927/3.73 
                  & 36.83/0.972/2.08
                  & 35.28/0.966/2.21      
                  & 37.10/0.977/1.52     
                  & 38.16/0.982/1.11   
                  & 36.95/0.978/1.18      
                  & 39.99/0.986/0.48
                  & 36.40/0.973/1.27     & \textbf{\textcolor{red}{40.04}}/\textbf{\textcolor{red}{0.988}}/\textbf{\textcolor{red}{0.43}}  \\
\textbf{Rain800}  & 21.97/0.827/6.25  & 23.62/0.835/5.79  & 22.41/0.838/5.73      & 26.61/0.831/4.88      & 26.72/0.865/4.61   & 26.32/0.874/4.47      & 27.33/0.869/4.54      & 27.50/0.876/4.22   & \textbf{\textcolor{red}{27.89}}/\textbf{\textcolor{red}{0.881}}/\textbf{\textcolor{red}{4.06}}    \\
\textbf{Rain100H} & 21.87/0.782/6.02 & 24.52/0.805/4.21 & 25.02/0.843/4.45      & 28.98/0.892/3.21      & 30.47/0.903/2.44   & 28.33/0.876/3.67      & 31.28/0.909/2.19      & 28.66/0.860/3.73     & \textbf{\textcolor{red}{31.44}}/\textbf{\textcolor{red}{0.911}}/\textbf{\textcolor{red}{2.07}}      \\
\bottomrule
\label{tab:quantitative_rain}
\end{tabular}}
\end{table*}
For the implementation detail, the initial learning rate is 10\ensuremath{^{-4}} and is multiplied by 0.75 after 10 epochs until the $250^{th}$ epoch. The Adam optimizer\unskip~\cite{554381:14619662} is applied, 16 directions and 4 levels are adopted in the CT. The proposed network is implemented on Nvidia RTX Titan GPU and trained with 1000
epochs. It takes thirty hours for training. Our network is trained based on PyTorch.
\subsection{Comparison with State-of-the-art Methods}\footnote[1]{Due to the limited space, more experimental results are reported in the supplementary material.}
We evaluate our method on two rain scenarios: moderate rain scenario and heavy rain scenario. For the moderate rain scenario, we apply several existing methods including DDN\unskip~\cite{fu2017removing}, JORDER\unskip~\cite{yang2017deep}, SPANet\unskip~\cite{wang2019spatial}, PreNet\unskip~\cite{ren2019progressive}, BRN\unskip~\cite{ren2020single}, DRDNet\unskip~\cite{deng2020detail}, RCDNet\unskip~\cite{wang2020model}, and MSPFN\unskip~\cite{jiang2020multi} for comparison. For fair evaluation, we ensure all models are trained with the same training set provided by the corresponding dataset.

\begin{figure*}[t!]
  \centering
  \subfigure[Synthesized \textbf{heavy rain} images\label{fig:HR-syn}]{\includegraphics[width=0.95\textwidth]{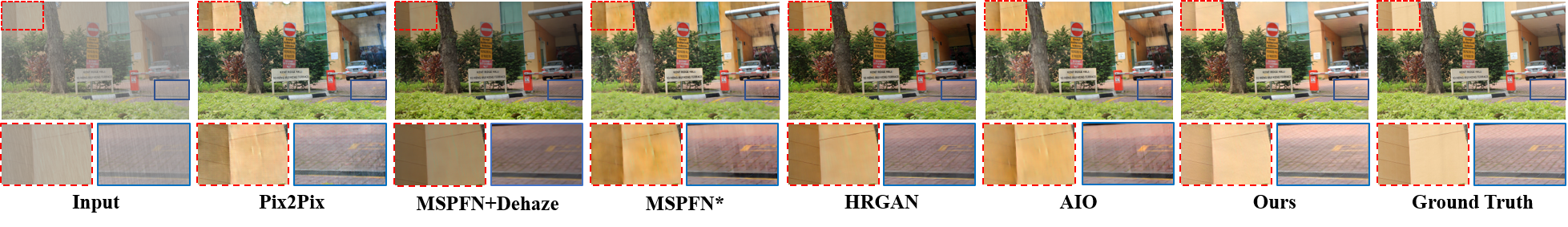}}
    \hspace{0.001em}
  \subfigure[Real-world \textbf{heavy rain} images\label{fig:HR-real}
]{\includegraphics[width=0.95\textwidth]{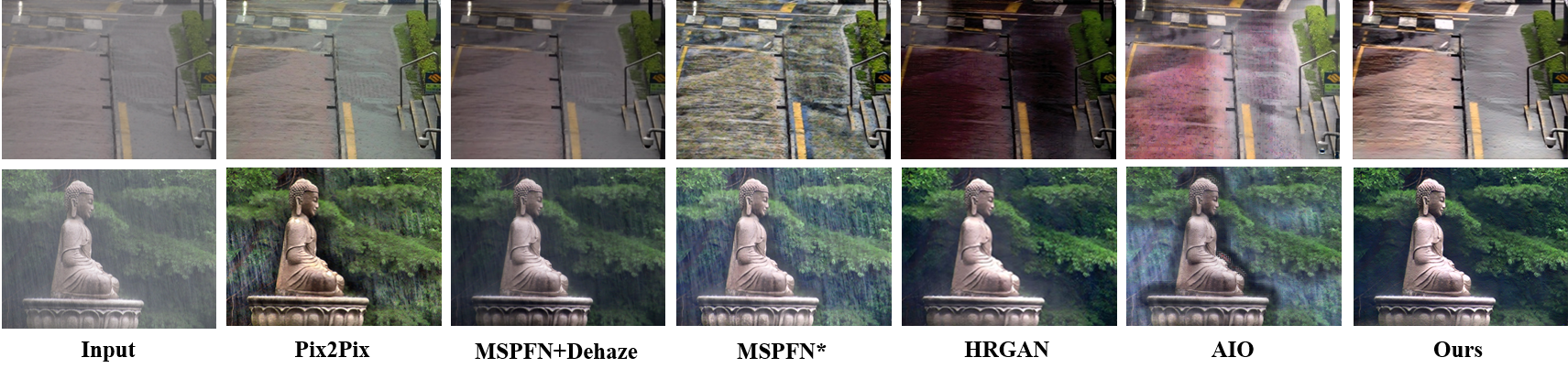}}
  \\  
  \subfigure[Synthesized \textbf{moderate rain} images.\label{fig:MR-syn}]{\includegraphics[width=0.95\textwidth]{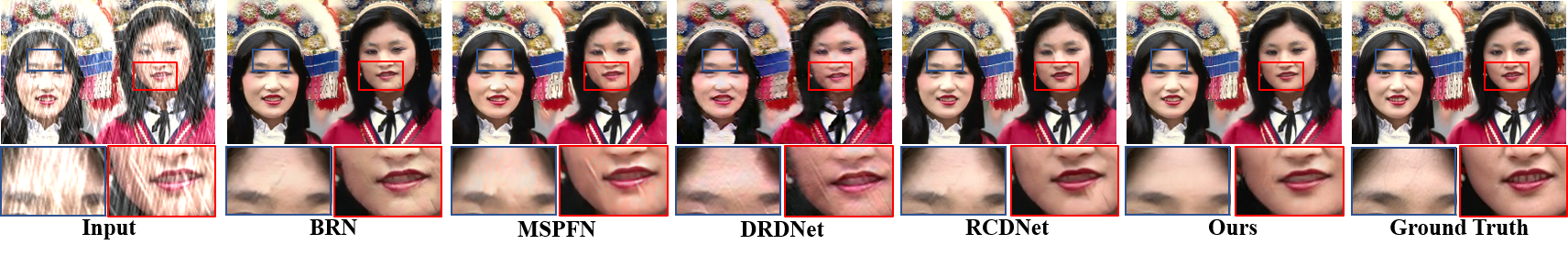}}
    \hspace{0.001em}
  \subfigure[Real-world \textbf{moderate rain} images\label{fig:MR-real}
]{\includegraphics[width=0.95\textwidth]{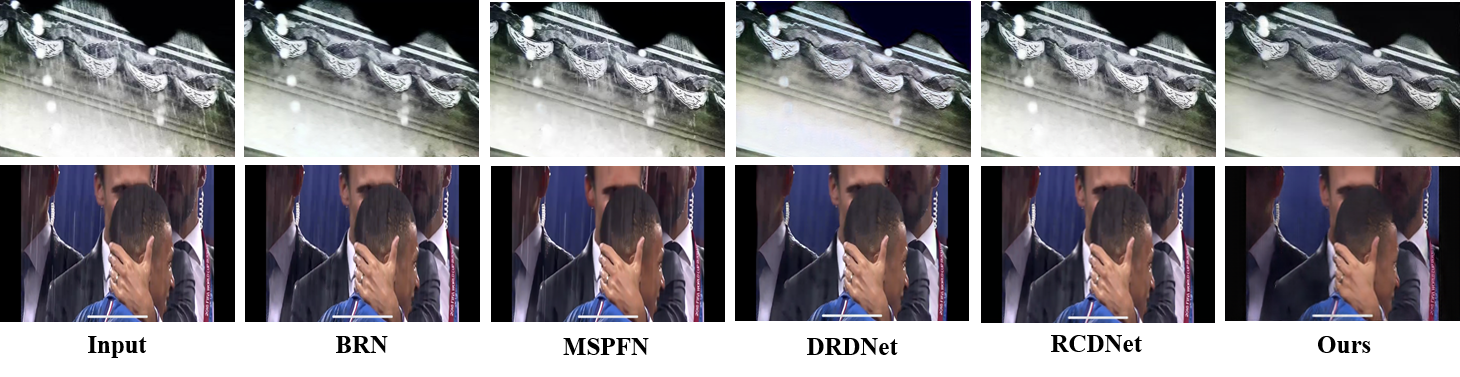}}
  \\  
  \caption{\textbf{Visual comparison with other state-of-the-arts algorithms for some examples in heavy rain scenario and moderate rain scenarios.}}
\label{fig:qualitative_HR}
\end{figure*}
For the heavy rain scenario, we adopt the HRGAN\unskip~\cite{li2019heavy} and AIO\unskip~\cite{li2020all} for comparison. Moreover, referring to the experiments in \unskip~\cite{li2019heavy}, we also compare our method with other heavy rain removal strategies: (i) moderate rain removal strategies combined with the dehazing strategy \unskip~\cite{Dong_2020_CVPR}; (ii) the moderate rain removal methods retrain with the heavy rain dataset; (iii) the image translation methods (the CycleGAN\unskip~\cite{zhu2017unpaired} and Pix2Pix\unskip~\cite{isola2017image}). For better identification, '*' denotes that the models retrain with the heavy rain dataset. Three metrics are applied for quantitative evaluation: the structural similarity (SSIM), the peak signal to noise ratio (PSNR), and the CIEDE 2000 color difference. A lower value of CIEDE2000 means less color distortion.
\noindent \smallskip\\
\textbf{Analysis on Heavy Rain Scene.} The quantitative comparisons are shown in \tabref{tab:quantitative_syn}. From \tabref{tab:quantitative_syn}, one can see that the proposed method outperforms other state-of-the-art heavy rain removal strategies in all metrics. Moreover, based on the visual comparison in Figures \ref{fig:HR-syn} and \ref{fig:HR-real}, one can see that, for existing heavy rain removal strategies, the recovered results tend to have residual rain streaks and color distortion. However, our method can solve these problems effectively and provide better visual quality compared to other methods.
\noindent \smallskip\\
\textbf{Analysis on Moderate Rain Scene.} 
\tabref{tab:quantitative_rain} presents the results of quantitative evaluation on conventional moderate rain datasets. One can see that the proposed method can achieve the best performance for moderate rain removal compared with other existing methods in all datasets and all metrics. Figures \ref{fig:MR-syn} and \ref{fig:MR-real} present the visual comparison on synthesized and real-world datasets. One can see that, compared with the proposed method, the results of existing methods may have more residual rain streaks (see the results presented in \figref{fig:MR-real}). These results show that the proposed method can achieve better rain removal and reconstruct the background effectively.

Based on the analysis above, the proposed method can achieve excellent performance on \textbf{both moderate rain and heavy rain scenarios}, which proves that the ContourletNet has the generalized ability on rain removal task.
 
\subsection{Ablation Study}
\begin{figure}[t!]
  \begin{minipage}[t]{0.4\textwidth}
    \centering
    \subfigure[Input]{\includegraphics[width=0.49\textwidth]{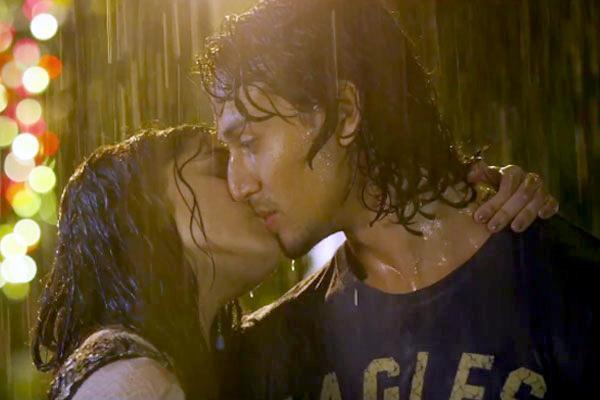}}
    \subfigure[MSPFN]{\includegraphics[width=0.49\textwidth]{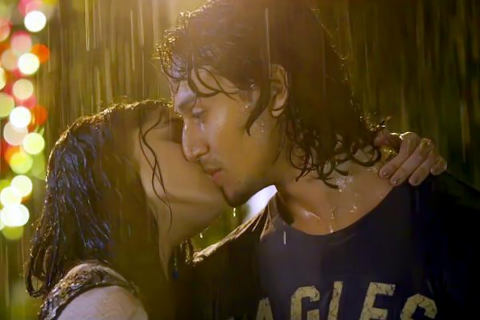}}
    \\
    \subfigure[RCDNet]{\includegraphics[width=0.49\textwidth]{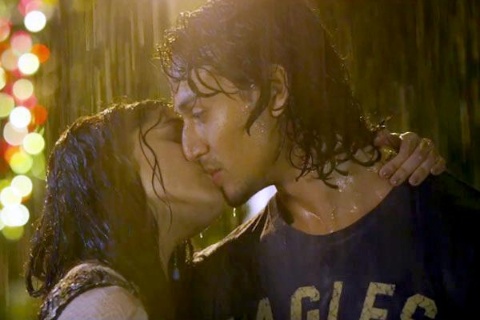}}
    \subfigure[Ours]{\includegraphics[width=0.49\textwidth]{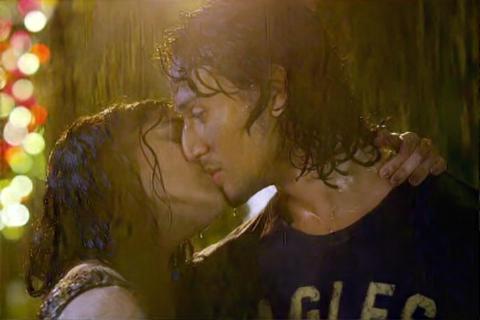}}
  \caption{Failure case of the proposed method and state-of-the-art rain removal methods under night-time rain scene.}
  \label{fig:fail-case}
  \end{minipage}
  \hfill
  \begin{minipage}[t]{0.58\textwidth}
    \captionof{table}{\textbf{Quantitative evaluation for ablation study.}}
        \centering
        \subtable[Effectiveness of the proposed modules]{
            \resizebox{\textwidth}{!}{
           \begin{tabular}{ccccccc} 
            \toprule
            \textbf{Dataset} & \textbf{C} & \textbf{C} \textit{w/o} \textbf{MD} & \textbf{C} \textit{w/o} \textbf{H} & \textbf{C} \textit{w/o} \textbf{G} & \textbf{C} \textit{w/o} \textbf{CP}& \textbf{C} \textit{w/o} \textbf{FEM} \\ 
            \hline\hline
            \textbf{Heavy} & \textbf{25.45}/\textbf{0.91} & 24.98/0.89 & 25.01/0.89 & 24.71/0.89 & 24.88/0.88 & 25.13/0.90 \\
            
            \textbf{Rain 100H} & \textbf{31.44}/\textbf{0.91} & 30.01/0.89 & 29.88/0.89 & 30.23/0.90 & 29.97/0.89& 30.41/0.90 \\
            \bottomrule
            \label{tab:ablation}
            \end{tabular}}
            }
        \\
        \subtable[Effectiveness of Contourlet Transform]{           
            \resizebox{\textwidth}{!}{
            \begin{tabular}{ccccccc} 
            \toprule
            \textbf{Dataset}   & \textbf{VConv} & \textbf{MConv} & \textbf{HL}& \textbf{LP} & \textbf{DWT} & \textbf{Ours}  \\
            \hline\hline
            \textbf{Heavy}     & 21.57/0.85     & 22.13/0.86     & 23.38/0.87 & 24.01/0.88  & 24.11/0.89   & \textbf{25.45}/\textbf{0.91}    \\
            \textbf{Rain 100H} & 26.21/0.84 & 27.14/0.85     & 29.31/0.87 & 29.51/0.88  & 29.63/0.88   & \textbf{31.44}/\textbf{0.91}    \\
            \bottomrule
            \label{tab:ablation_differentFE}
            \end{tabular}
            }          
            }
        \subtable[Level of CT decomposition versus the performance of recovery.]{           
            \resizebox{\textwidth}{!}{
            \begin{tabular}{ccccccc} 
            \toprule
            \textbf{Dataset} & \textbf{Lv 1} & \textbf{Lv 2} & \textbf{Lv 3} & \textbf{Lv 4}  & \textbf{Lv 5}  & \textbf{Lv 6} \\ 
            \hline\hline
            \textbf{Heavy} & 21.95/0.86 & 23.88/0.88 & 24.79/0.90 & 25.45/0.91 & 25.48/0.91 & 25.50/0.91 \\
            \textbf{Rain100H} & 29.69/0.87 & 30.18/0.88 & 30.72/0.90 & 31.44/0.91 & 31.47/0.91 & 31.48/0.91 \\
            \bottomrule
            \label{tab:levels}
            \end{tabular}
            }}
    \label{tab:multi_ablation}
    \end{minipage}
  \end{figure}

To verify the effectiveness of each of the proposed modules in this paper, five combinations are performed on the heavy rain and Rain 100H datasets: (1) the proposed ContourletNet with multi-level discriminator (\textbf{C}) (2) the ContourletNet with single-level discriminator (\textbf{C} \textit{w/o} \textbf{MD}); (2) \textbf{C} without the hierarchical architecture (\textbf{C} \textit{w/o} \textbf{H}); (3) \textbf{C} without the aggregate contourlet component (\textbf{C} \textit{w/o} \textbf{G}); (4) \textbf{C} without the contourlet predictor (\textbf{C} \textit{w/o} \textbf{CP}); (5) \textbf{C} without the feedback error map mechanism (\textbf{C} \textit{w/o} \textbf{FEM}). \tabref{tab:ablation} shows that the best performance can be achieved if all proposed techniques (the hierarchical architecture, the aggregate contourlet component, the contourlet predictor, the feedback of error map, and the multi-level contourlet components discriminator) are adopted. 

To prove the effectiveness of the CT, we apply several existing feature extraction techniques instead of the CT in the proposed network for comparison: the vanilla convolution operation with a 3$\times$3 kernel (\textbf{VConv}); multi-scale convolution kernel (\textbf{Mconv})\unskip~\cite{ren2016single}; Laplacian pyramid (\textbf{LP})\unskip~\cite{fu2019lightweight}; HL\unskip~\cite{li2019heavy}; DWT\unskip~\cite{554381:14618851}. The results in \tabref{tab:ablation_differentFE} indicate that using the CT as the rain streak extractions can achieve the best result on deraining. Moreover, we present a visual comparison on the real-world scene in the Supplementary Material.

In \tabref{tab:levels}, we discuss the effect of decomposition levels for the CT and the recovered performance. One can see that increasing the levels of decomposition can benefit the reconstruction quality generally. The more levels of decomposition, the better performance the proposed method may have.

\section{Conclusion}
In this paper, to mitigate the limitation in existing rain removal methods, a novel ContourletNet is proposed. First, to address the residual rain streaks in the recovered image, we leverage the CT to decompose the input image to several multi-direction subbands and a semantic subband at each level. Second, based on observing
 the semantic subbands of rainy images and their corresponding ground truths, the hierarchical recovery process is proposed. Last, the multi-level subband discriminator is proposed to suppress the error propagation in each subband. Extensive experiments prove the effectiveness of the proposed ContourletNet for both heavy and moderate rain scenarios. In the future work, we will address the limitation of the proposed method as shown in \figref{fig:fail-case}. Specifically, the proposed method may fail for night-time rainy scene. There still exists residual rain streaks in the recovered images.
 
\section{Acknowledgement}
We thank to National Center for High-performance Computing (NCHC) for providing computational and storage resources.

\bibliographystyle{IEEE}
\bibliography{egbib}

\end{document}